\pdfoutput=1

\documentclass[11pt]{article}

\usepackage[final]{acl}

\usepackage{times}
\usepackage{latexsym}

\usepackage[T1]{fontenc}

\usepackage[utf8]{inputenc}

\usepackage{microtype}

\usepackage{inconsolata}

\usepackage{graphicx}
\usepackage{pgf}
\usepackage{xcolor}
\usepackage{tikz}
\usetikzlibrary{positioning,fit,backgrounds,shapes,shadows.blur,arrows.meta, shadows, fit, calc}
\usepackage{hyperref}
\usepackage{subcaption}
\usepackage{enumitem}
\usepackage{booktabs}
\usepackage{multirow}

\title{The Anatomy of Evidence: An Investigation Into Explainable ICD Coding}

\author{
 \textbf{Katharina Beckh\textsuperscript{1,2}},
 \textbf{Elisa Studeny\textsuperscript{1}},
 \textbf{Sujan Sai Gannamaneni\textsuperscript{1,2}},
\\
 \textbf{Dario Antweiler\textsuperscript{1}},
 \textbf{Stefan Rüping\textsuperscript{1}}
\\
\\
 \textsuperscript{1}Fraunhofer IAIS 
 \textsuperscript{2}Lamarr Institute
\\ 
 \small{ \texttt{
    katharina.beckh@iais.fraunhofer.de}
 }
}

\begin{document}
\maketitle

\begin{abstract}
Automatic medical coding has the potential to ease documentation and billing processes. For this task, transparency plays an important role for medical coders and regulatory bodies, which can be achieved using explainability methods. However, the evaluation of these approaches has been mostly limited to short text and binary settings due to a scarcity of annotated data. Recent efforts by \citet{cheng-etal-2023-mdace} have introduced the MDACE dataset, which provides a valuable resource containing code evidence in clinical records. In this work, we conduct an in-depth analysis of the MDACE dataset and perform plausibility evaluation of current explainable medical coding systems from an applied perspective. With this, we contribute to a deeper understanding of automatic medical coding and evidence extraction. Our findings reveal that ground truth evidence aligns with code descriptions to a certain degree. An investigation into state-of-the-art approaches shows a high overlap with ground truth evidence. We propose match measures and highlight success and failure cases. Based on our findings, we provide recommendations for developing and evaluating explainable medical coding systems.
\end{abstract}

\section{Introduction}

Explaining model predictions is a key component in automatic medical coding systems. 
Healthcare systems worldwide employ coding systems to document diagnoses and procedures, and to process medical bills.
Codes are often structured in hierarchies which contain up to hundreds of thousands of terms to cover the complexity of medical care and the multitude of synonymous medical expressions. 
Medical codes are used for administrative purposes \cite{klug2024}, for prediction of length of stay and mortality \cite{Harerimana2021}, adverse event detection, and clinical decision support \cite{Antweiler2023}.
The \emph{International Classification of Diseases} (ICD)\footnote{\url{https://www.who.int/standards/classifications/classification-of-diseases}} is the most prominent system in hospital settings.  
A patient's set of ICD codes together with the main diagnosis dictates the level of reimbursement. 
Assigning appropriate ICD codes manually requires extensive expertise and is guided by detailed, country-specific rule books.

\begin{figure}
\centering
\begin{tikzpicture}[
    auto,
    font=\sffamily,
    document/.style={
        rectangle,
        rounded corners,
        draw=none,
        fill=white,
        text width=7.0cm,
        align=justify,
        drop shadow={shadow blur steps=2},
    },
    note/.style={
        rectangle,
        draw=none,
        fill=blue!20,
        inner sep=3pt,
    },
    arrow/.style={
        -{Latex[length=3mm]},
        thick,
    }
]
\definecolor{blue_new}{HTML}{648FFF}
\definecolor{yellow_new}{HTML}{FFB000}
\node [document] (document) {
\begin{minipage}{\textwidth}
\small
The patient was admitted with severe 
\tikz[baseline={(hypertension.base)}] \node [fill=blue_new!30, rounded corners,anchor=base] (hypertension) {hypertension};\tikz[baseline={(note1.base)}] \node [fill=blue_new!70, rounded corners,anchor=base] (note1) {I10};.
Beta-blockers and ACE inhibitors were administered. The patient reported a\tikz[baseline={(smoking.base)}] \node [fill=yellow_new!20, rounded corners,anchor=base] (smoking) {history of smoking};\tikz[baseline={(note3.base)}] \node [fill=yellow_new!80, rounded corners,anchor=base] (note3) {F17};
and high cholesterol levels.
\end{minipage}
};



\end{tikzpicture}
\caption{Discharge summary with ICD codes and respective evidence highlighted.}
\end{figure}

To reduce manual efforts and increase efficiency, automatic coding methods have been developed. 
With the adoption of deep learning, especially the transformer architecture \cite{vaswani2017,devlin-etal-2019-bert}, automatic coding systems have achieved practical usage \cite{Biswas2021, edin2024}. State-of-the-art approaches treat code prediction as a multi-label classification task and rely on pre-trained language models \cite{ji2024unified}.    
However, these models with billions of parameters lack transparency -- which diminishes acceptance by coding staff and presents a barrier during regulatory assessments by external authorities.

The goal is to develop medical coding systems which are able to provide text evidence for a predicted code. 
This objective can be achieved with the use of explainability methods.  
In particular, feature attribution methods identify which input features are decisive for the model output \cite{kim-etal-2022-current, atanasova-etal-2020-diagnostic, ribeiro2016, lundberg2017} assigning scores to each token of a medical document.   
However, explainability research has been mostly constrained to short text and binary classification on social media data or product reviews \cite{wiegreffe2021, guzman2024}. 
In the medical domain, the scarcity of data and the need for domain expertise make word level annotations challenging.
\citet{cheng-etal-2023-mdace} carried out a re-annotation effort of a MIMIC-III subset \cite{johnson2016mimic} and created annotation spans as evidence for each medical code. 
This dataset is the first to include textual evidence for assigned medical codes and offers new research opportunities in information extraction and explainability. 

Building on this, recent work investigated different training strategies and evidence extraction methods showing promising results using gradient-based explanations \cite{edin2024}.\footnote{Throughout the paper, evidence and explanation are used interchangeably.} The evaluation focused on two key properties of explanations, faithfulness and plausibility \cite{jacovi-goldberg-2020-towards, nauta2023anecdotal}. However, prior work was technique-centric, lacking depth in data analysis and application-focused evaluation. 

Hence, what is still missing is an investigation from a practical perspective on (a) how to utilize the dataset for research purposes and (b) how well current explainability approaches, incl. evaluation, work for clinical adoption.   

This work aims to deepen the understanding of ICD coding, and improve evidence extraction and evaluation.\footnote{Our code is available at \url{https://github.com/lamarr-xai-group/anatomy-of-evidence}} The focus is on data understanding and plausibility evaluation. 
Our contributions can be summarized as follows:
\begin{enumerate}
\itemsep0em 
    \item We carry out an in-depth analysis of the MDACE dataset and provide insights for dataset utilization. 
    \item We perform experiments with state-of-the-art explainable medical coding systems, introduce match measures and provide a practical perspective on results. 
    \item Based on our findings, we give actionable recommendations for evidence extraction and evaluation in ICD coding.
\end{enumerate}


\section{Dataset Overview}

MDACE is a medical dataset in the English language containing Electronic Health Records (EHRs) along with associated ICD codes \cite{cheng-etal-2023-mdace}. The EHRs included in MDACE are a subset of the larger MIMIC-III dataset \cite{johnson2016mimic, goldberger2000physiobank} containing 302 admissions.\footnote{Use of the data and trained models is restricted to scientific research.} They have been recoded from ICD-9 to ICD-10 and extended by text spans as evidence for these respective codes. This process was carried out by two teams of professional medical coders. One team (Inpatient) coded all documents and annotated in a \textit{sufficient} manner, i.e., as much evidence as necessary to justify the respective code, while the other team (Profee) coded a subset in a \textit{complete} manner, i.e. all evidence contained within the documents. 

\begin{figure}
    \begin{center}
    \hspace*{-10pt}\input{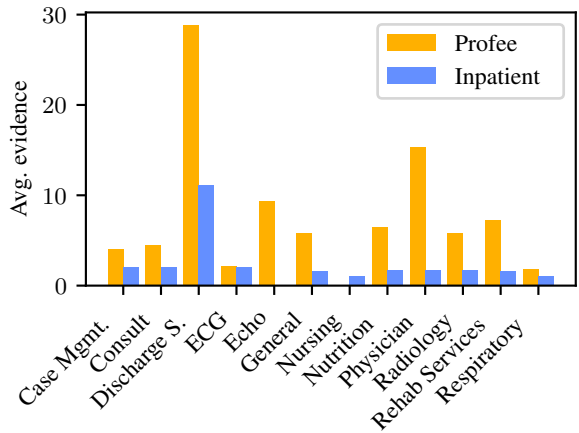}
    \end{center}
    \caption{Average number of evidence spans for each document type for Profee and Inpatient}
    \label{fig:amount_per_doc}
\end{figure}

\paragraph{Amount and length}
Across all admissions, documents and codes, the dataset contains 9,499 evidence spans, 5,563 in Profee and 3,936 in Inpatient. 
The average labels per document is 11.3 for Inpatient and 31.4 for Profee. Hence, there is roughly three times more evidence in the complete labeling scheme. 
\autoref{fig:amount_per_doc} shows the average number of evidence  per document type for Inpatient and Profee. 
Evidence length for Inpatient and Profee is roughly the same, 2.18 and 1.96 tokens.

\section{Analysis of Human-Annotated Evidence}
To determine the characteristics of medical code evidence, we perform an in-depth analysis of the MDACE dataset.  
Compared to other data resources, the documents in MDACE are relatively long with an average of around 2,000 words \cite{dai-etal-2022-revisiting}. 
Given the considerable length, we explore whether focusing on relevant segments could be a driver for computational efficiency. 
To the same effect, we seek efficient methods to detect relevant labels. We analyze to what extent evidence and medical codes can already be captured with existing knowledge, i.e., code descriptions.
Prior work did not discuss the relation of the common documents in Inpatient and Profee which we address in this work.  
In a qualitative analysis, we shed light on annotation dependencies and annotation diversity. 
These points give rise to the following research questions:
\begin{itemize}
    \item \textbf{\texttt{RQ1}} Where in the document is ground truth evidence located?
    \item \textbf{\texttt{RQ2}} Does ground truth evidence overlap with code descriptions?
    \item \textbf{\texttt{RQ3}} Is sufficient evidence a subset of complete evidence?
\end{itemize}

\subsection{RQ1: Where in the document is ground truth evidence located?}

\begin{figure}
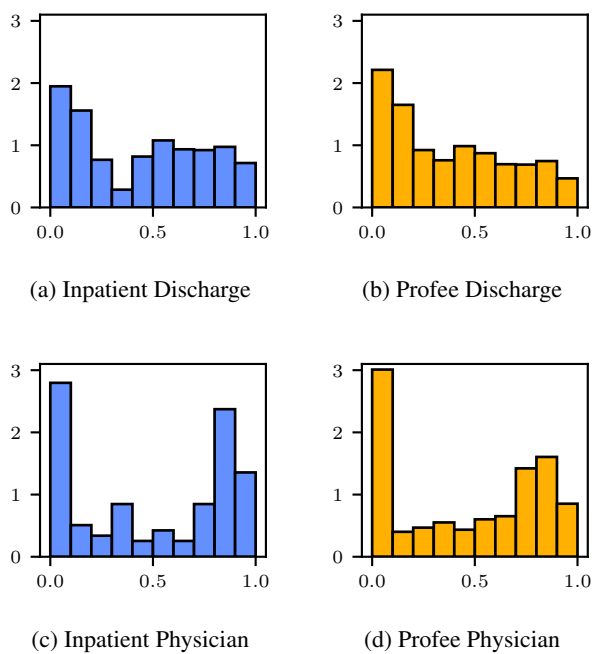

    \centering
    \begin{subfigure}{0.45\columnwidth}
        \centering
        \hspace*{-7pt}\input{figures/position_i_discharge.pgf}
        \caption{Inpatient Discharge}
        \label{fig:subfig1}
    \end{subfigure}
    \hfill
    \begin{subfigure}{0.45\columnwidth}
        \centering
        \hspace*{-7pt}\input{figures/position_p_discharge.pgf}
        \caption{Profee Discharge}
        \label{fig:subfig2}
    \end{subfigure}
    \vskip\baselineskip
    \begin{subfigure}{0.45\columnwidth}
        \centering
        \hspace*{-7pt}\input{figures/position_i_physician.pgf}
        \caption{Inpatient Physician}
        \label{fig:subfig3}
    \end{subfigure}
    \hfill
    \begin{subfigure}{0.45\columnwidth}
        \centering
        \hspace*{-7pt}\input{figures/position_p_physician.pgf}
        \caption{Profee Physician}
        \label{fig:subfig4}
    \end{subfigure}
    
    \caption{Distribution of human-annotated evidence positions based on a common subset annotated by both Inpatient and Profee coders. The four subfigures depict the distributions for  discharge summaries and physician notes for Inpatient and Profee annotations. The x axis refers to relative evidence positions and the y axis shows normalized evidence counts.}
    \label{fig:position}
\end{figure}

For the position analysis, we examine whether focusing on specific text segments, e.g., beginning of a document, is beneficial for solving the classification and evidence extraction tasks.   

We concentrate on the two most frequent and commonly used document types: Discharge summaries and physician notes. 
The underlying data basis is the common subset of documents annotated by both, Profee and Inpatient coders.
\autoref{fig:position} shows the distribution of evidence position throughout the discharge summaries and physician notes for Inpatient and Profee. Evidence counts are normalized per plot and a relative position of 0.5 means that evidence appears halfway through the document.
For Inpatient, evidence occurs mostly at the beginning and end of documents. 
In comparison, the relevant information in physician notes seems to occur more towards the end of documents. 
For Profee, a valley is only perceivable for the physician notes, there seems to be no such pattern in the discharge summaries. 

\paragraph{Does sufficient evidence occur earlier in the document?}
Since text is typically scanned from top to bottom, we assumed that sufficient evidence is annotated when it first occurs in the document. That leads to the hypothesis that sufficient evidence occurs more in the beginning of documents in comparison to complete evidence. 
The position information shows that this is not the case: The evidence counts in Inpatient are not higher for lower positions.  
Contrary to our assumption, sufficient evidence is not located more in the beginning of the documents.  

\subsection{RQ2: Does ground truth evidence overlap with code descriptions?}
The aim behind this question was to derive insight from human-annotated evidence to inform development of efficient methods for the ICD coding task. 
If we can find evidence and identify ICD codes with existing knowledge, computing resources may be reduced. 
As knowledge base, we use code descriptions -- human-readable titles for the alphanumeric ICD codes available in the ICD system. For example, the code description for 427.31 is `Atrial fibrillation'. 
We investigate to what extent code descriptions overlap with human-annotated evidence, in turn, giving an indication of whether certain codes can be detected with code descriptions.

\begin{figure}
    \centering
    \hspace*{-10pt}\input{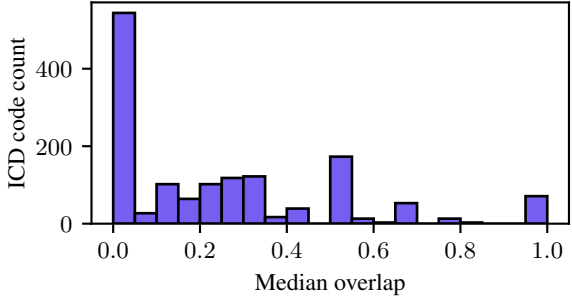}
    \caption{Distribution of ICD codes according to evidence overlap with code descriptions. The x axis shows the median value of relative overlap scores, the y axis refers to the code counts.}
    \label{fig:description_evidence}
\end{figure}

For this analysis, the ICD-10-CM system and code descriptions of the MDACE dataset were used.
The evidence spans and code descriptions were lemmatized; stop words and punctuation were removed. For example, the code description of R06.83 `snoring’ is lemmatized to `snore’. For each code the intersection of evidence-description pairs is calculated and divided by the set of words in the description. The median for all spans of that code is used to display the overlap. \autoref{fig:description_evidence} shows that there are roughly three overlap categories: nearly no overlap, overlap to some extent and strong overlap. Most codes fall into the no or nearly no overlap category as the overlap is close to zero. 
However, a number of codes have a strong overlap with the code description. For example, two codes with high overlap are R06.83 (snoring) and also I31.3 (pericardial effusion).
Codes with high overlap of evidence and code descriptions present research potential for solving part of evidence extraction with efficient methods such as rule-based systems.

\subsection{RQ3: Is sufficient evidence a subset of complete evidence?}
One unique feature of the MDACE dataset is the annotation in both, a sufficient (Inpatient) and complete manner (Profee).
The goal was to establish the relationship between Inpatient and Profee annotations.
With the hypothesis that sufficient evidence is a subset of the complete documents, we selected the 52 document with common hospital admission ID, compared the codes and calculated a strict subset.   
From the note IDs it became clear that the common subset is smaller than anticipated with 470 unique note IDs and only 118 common note IDs. 
Of these 118 common note IDs 1602 codes are unique and 331 are identical. 
From these 331, 55 are a subset. 
\citet{cheng-etal-2023-mdace} state that ICD codes do not necessarily align due to different coding rules. However, the alignment is still smaller than anticipated.

\subsection{Qualitative Analysis}

\begin{table}
    \small
    \centering
    \caption{Two ICD codes with information on corresponding evidence: number of occurrences in the whole data set / unique counts, and evidence examples}
    \begin{tabular}{llp{3.8cm}}
        \toprule
        \textbf{Code} & \textbf{all/unique}& \textbf{Evidence examples} \\ \midrule
        I10     & 133/8 & \small{`hypertension', `HTN', `hypertensive`} \\
        Z87.891  &  20/19 &  \small{`smoking history',  `former smoker', `the distant past'}\\
        \bottomrule
    \end{tabular}
    \label{tab:diversity}
\end{table}

\paragraph{Evidence diversity per code}
We examined the linguistic diversity of evidence within codes. For each code, the corresponding evidence spans were compiled across all document types. For the majority of codes, the evidence appears similar, e.g., as for ICD code I10 (Essential (primary) hypertension) shown in \autoref{tab:diversity}. Evidence for this code occurs frequently in the data, whereas the number of unique strings is small. In contrast, evidence for code Z87.891 (Personal history of nicotine dependence) is more diverse; the unique count nearly equals the number of occurrences. 
This finding suggests that some codes have distinct evidence while others present with higher variance which may make them more difficult to learn. 

\paragraph{Dependencies between evidence spans}
When performing sanity checks, we found evidence spans that would not justify a code per se. Taking the example of Z81.891 with paraphrased context: \textit{`Smoked one PPD for seven years in the distant past'} with `in the distant past' as evidence. Here, `in the distant past' is not sufficiently capturing both the notion of past and smoking.  
We hypothesize that this may be due to two factors: (1) A link is missing, that is, the reference or relation to another annotation for the same code is missing and  (2) underspecification, i.e., information is not annotated explicitly enough, which can be the result of missing information due to anonymization or annotation oversight.

\section{Analysis of Model Explanations}
For the modeling approaches, we investigate to what extent current approaches for explainable ICD coding are able to extract plausible evidence that matches with the ground truth. 
In a qualitative analysis, we highlight success and failure cases. 
We were interested in finding out whether model correctness relates with explanation characteristics, in particular explanation length. 
Taken together, we consider the following research questions  for the analysis of model evidence: 

\begin{itemize}
    \item \textbf{\texttt{RQ4}} How does the length of explanations relate to classification performance?
    \item \textbf{\texttt{RQ5}} How well do model explanations match with ground truth explanations?
    \item \textbf{\texttt{RQ6}} To what extent do different model approaches match w.r.t. evidence?
\end{itemize}

\subsection{Experimental Setup}
\paragraph{Data split}
The data consists of discharge summaries from the Inpatient charts (sufficient). Following prior work, the ICD-9 system is used. 
The split for training, validation and test is 181/60/61 \cite{cheng-etal-2023-mdace}.
The 61 documents in the test set contain 586 evidence spans.
Notably, the evaluation includes true positives and false negatives, i.e., ICD codes that were not predicted are nevertheless part of the explanation evaluation.

\paragraph{Models}
The analysis is performed on state-of-the-art fine-tuned models available from \citet{edin2024}.
There are five model types -- each underwent a different training strategy. The first type relies on a supervised approach which uses annotated evidence in the training objective minimizing the Kullback-Leibler divergence between the cross-attention weights and the explanation spans \cite{huang-etal-2022-plm, edin2024}.   
The second type is an unsupervised baseline, i.e., no evidence annotations were used during training. The other three types are also unsupervised but employ robustness strategies to reduce relative importance of irrelevant tokens. The strategies are input gradient regularization, projected gradient descent and token masking as detailed in \cite{edin2024}. 
For each type, 10 trained models are available based on different seeds resulting in 50 model weights. 

For \textbf{\texttt{RQ4}}, all 50 models are utilized to investigate how evidence length relates to classification performance overall. 
For \textbf{\texttt{RQ5}} and \textbf{\texttt{RQ6}}, we focus on one supervised and one unsupervised model to dive deeper into evidence extraction from a practitioner's perspective. A typical applied setting is assumed where the best performing models are selected for integration and deployment. 
The two model seeds are selected according to explanation metrics. The best-performing unsupervised models is a model with input gradient regularization (IGR).\footnote{Seed numbers for best-performing models - unsupervised (IGR): igr/1p0vue7o and supervised: supervised/r5u1sr8h} This approach regularizes with the L2 norm of the gradient of the loss w.r.t. the input token embedding. 

The models are based on a modified version of the PLM-ICD architecture \cite{huang-etal-2022-plm}; the label-wise attention has been replaced by standard cross-attention to improve stability \cite{edin2024}. The underlying model is a RoBERTa architecture \cite{liu2019robertarobustlyoptimizedbert} pre-trained on PubMed text and physician notes \cite{lewis-etal-2020-pretrained}. Fine-tuning was performed on MIMIC-III discharge summaries, utilizing MDACE evidence spans for the supervised strategy.

\paragraph{Explanation Method}
For the generation of evidence, we used the feature attribution method AttInGrad which achieved the best performance according to faithfulness and plausibility metrics when compared to other gradient- and perturbation-based methods in prior work \cite{edin2024}. AttInGrad is a combination of Attention and Input$\times$Grad, where the attention weights are multiplied by the feature attribution scores attained by applying the L2 norm to the Input$\times$Grad feature attributions. For Input$\times$Grad, the input as well as the gradients are multiplied element wise.
A decision threshold for the attribution scores is calculated and set based on the validation set.

\paragraph{Match measures}
Previous work evaluated plausibility of explanations with the F1 score or Intersection-Over-Union \cite{deyoung-etal-2020-eraser}. 
These metrics are suitable for benchmarking, however, they are less meaningful for answering how close model explanations are to the ground truth. The underlying assumption is that it is often enough to guide the attention of the coder to a context window. 
We therefore introduce measures which are more comprehensible for an end user and which capture this notion of proximity. 
For the comparison of ground truth evidence and model evidence we use the following measures:  

\begin{description}
    \itemsep0em 
    \item \textbf{Empty}: No evidence is predicted because the attribution scores are below the threshold for inclusion.
    \item \textbf{Exact match}: The set of ground truth evidence is equal to the set of model evidence.
    \item \textbf{Proximate match}: All ground truth sequences have at least one token match in the set of machine explanations and tokens that have no match are in a context window of \( k \).\footnote{\( k \) is set to 10}
    \item \textbf{Partial match}: At least one ground truth sequence does not have a match or at least one machine explanation token is outside the context window.
    \item \textbf{No match}: There is no overlap in token IDs.
\end{description}

\subsection{RQ4: How does the length of explanations relate to classification performance?}

\begin{figure}
    \centering
    \hspace*{-10pt}\input{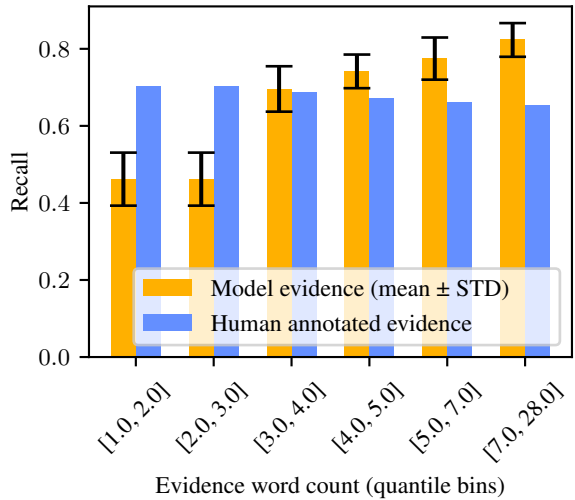}
    \caption{Recall with standard deviations of ICD codes in relation to evidence word count. Comparison of explanations from all models with ground truth (GT) explanations. Bins correspond to evidence word count.}
    \label{fig:ev_perform}
\end{figure}
Prior work found that models are better at extracting evidence with low word count \cite{cheng-etal-2023-mdace}. While this is a characteristic for evidence extraction abilities, is short model evidence also an indicator for ICD classification performance?    
We explore how evidence length, measured by the number of words, relates to recall. 
In~\autoref{fig:ev_perform}, the relation of evidence length and ICD code recall is depicted for all 50 model seeds. In this plot, the blue bars display the recall of all models when grouped over word count of the human-annotated evidence (in quantile bins). While only minor variations are observed here, strong impact on recall is observed for model evidence word count (in quantile bins) shown in orange bars. In these cases, when fewer words are extracted as evidence, the models are more likely to be wrong than when more words are predicted as evidence.

\subsection{RQ5: How well do model explanations match with ground truth explanations?}

\begin{figure}
    \hspace*{-10pt}\input{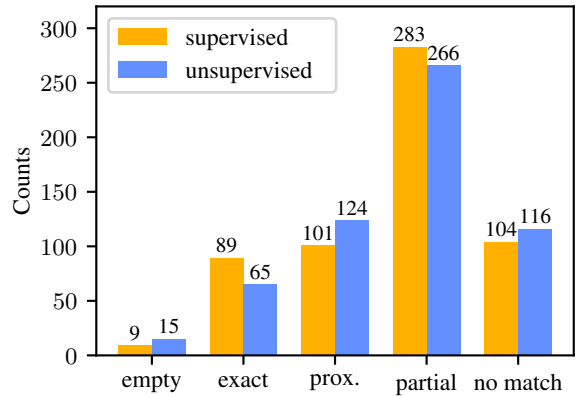}

    \caption{Comparison of the supervised and unsupervised model regarding evidence match on the test set showing the numbers of empty, exact match, proximate match and no match. Lower empty, lower no match and higher exact match is desired.}
    \label{fig:counts_simple}
\end{figure}

The heart of evaluating plausibility lies in how good medical coding systems are in providing evidence. 
We investigate to which extent the extracted ICD evidence from models aligns with human-annotated evidence to determine similarities, differences, and potential for automation.
\autoref{fig:counts_simple} shows match counts for the supervised and the unsupervised model on the test cases. 

\paragraph{Overlap of human-annotated and model evidence is high.} In roughly 80\% of test cases, the best model (supervised) identified at least one correct token (exact, proximate and partial match). The number of exact matches are low in comparison to the overlap counts. The cases for which no explanation is provided, due to scores being below explanation threshold, is minimal. 

\paragraph{No id match can still be a semantically meaningful match} 
In an effort to understand whether the no match category contains semantically meaningful evidence, all evidence in that category was manually annotated by two authors with basic medical knowledge. A binary label was assigned to signify the semantic match of the respective strings. The context was not considered, only the evidence strings. A strict approach was taken, only assigning a positive label when each token in the ground truth has an equivalent string-match or close match such as `obesity' and `obese'.
We find that roughly half of the non-matches align with the ground truth evidence (46\% for supervised and 53\% for unsupervised).
For a more precise analysis, expert assesments are necessary.

\subsection{RQ6: To what extent do the model approaches agree w.r.t. evidence?}

\begin{figure}
    \input{figures/heatmap1.pgf}
    \caption{Alignment of match types between supervised and unsupervised model based on the five match types using counts.}
    \label{fig:heatmap}
\end{figure}

When evaluating explanations, the Rashomon effect has to be taken into account. This effect implies that models can have equal performance but different underlying strategies \cite{mueller2023}. Therefore, we compare the evidence from different models, unsupervised and supervised, to gain insight into explanation agreement.  

Agreement, in terms of evidence, is assessed based on the match types (see \autoref{fig:heatmap}). 
In 74\% of cases the models have the same match type (diagonal values). For exact matches, they agree on 49 cases. 
The supervised model has fewer empty evidence predictions and fewer non-matches, which is desired. The number of exact matches and partial matches are higher, while proximate matches are lower.
Cases assigned an exact match by the supervised model but not by the unsupervised model received a proximate or partial match, with only two cases as non-matches.

\paragraph{Probabilities}
Considering code predictions, all codes that have at least an output probability of 0.5 are treated as predicted. Taking a closer look at the performance on a code-level:
For the supervised model, 374 of 586 cases are true positives (TP) and the rest are false negatives (FN), while the unsupervised model has 379 out of 586 TP. 
The number of TP is similar for both models and the number of FN is relatively high, but within expectation given the large label space. 
In addition, \autoref{tab:prob} reports the average probability and standard deviations for each match count and model type. 
The probabilities for empty evidence predictions are low for both models, especially for the supervised approach. For inference, empty evidence is therefore rare and negligible. 
Exact matches are not tied to higher probability values. 

\begin{table}
    \centering
    \caption{Average probabilities of code predictions for supervised and unsupervised model sorted by match types, standard deviation in parentheses}
    \begin{tabular}{lcc}
        \toprule
	\multirow{2}{*}{\textbf{Type}}&\multicolumn{2}{c}{\textbf{Average probability}}\\
         & supervised & unsupervised  \\
        \midrule
        empty & 0.002 ($\pm$0.006) & 0.102 ($\pm$0.226)\\
        exact & 0.607  ($\pm$0.300) & 0.637 ($\pm$0.296)\\
        prox & 0.496  ($\pm$0.349) & 0.561 ($\pm$0.325)\\
        partial & 0.688  ($\pm$0.310) & 0.707 ($\pm$0.303)\\
        no match & 0.471 ($\pm$0.373) & 0.467 ($\pm$0.386)\\
        \bottomrule
    \end{tabular}
    \label{tab:prob}
\end{table}

\subsection{Qualitative Analysis}
\paragraph{Do exact matches have a pattern?}
The observation across supervised and unsupervised is that exact matches are often well-defined words: 
Anemia, GERD, Alzheimer’s disease, osteoarthritis, hypertension, etc.
However, the presence of these words is not a sufficient condition for an exact match, i.e., just the occurrence of `hypertension' does not automatically lead to an exact match.  

\paragraph{Abbreviations and uppercasing}
We initially anticipated that the models would struggle with abbreviations. This did not prove to be the case as there are many success cases in the test data, e.g., `GERD' and `Gastroesophageal reflux' are both extracted correctly.
However, uppercasing seems to present a challenge. For example, 
`HYPERBILIRUBINEMIA' gets tokenized into: HYP, ER, BI, LI, R, UB, IN, EM, IA. The two models did not identify the relevant tokens. 
If it is known that uppercasing occurs frequently, a preprocessing step may be beneficial. 
Furthermore, the model evidence contains punctuation to a small degree, but the occurrence rate is relatively low. One option is to remove punctuation before tokenization, although this cleaning step is not advisable because information may get lost, e.g., for `C. difficile'. 

\paragraph{Duplicates}
Models tend to extract more evidence than humans annotated, often echoing the same or similar information at different positions.
For example `hypertension' in four different locations in the text. 
If duplicates are undesired, a post-processing step could cushion the effect. 

\paragraph{Confused or correct?}
We found unexpected correlations in the model evidence. 
In certain cases, the models did not only extract a condition but also a drug as evidence, e.g., `hypertension' and `ACE' or `shingles' and `acyclovir'.
This leads to the question whether drugs count as evidence. On the one hand, this can be seen as an undesired spurious correlation because the drug may not be a sufficient condition for a diagnosis, considering that drugs can be repurposed. 
If the drug was taken before the hospital stay, it is an indicator of a disease, but it should not be the main explanation. If the medication was administered after the diagnosis, it has no explanatory power. 

Similarly, there are signs of co-morbidities. For example, ICD code 311 (depressive disorder) received `depression' as well as `anxiety' as evidence. 
Taken together, these correlations again demonstrate how feature attribution does not produce causal relations.

\section{Discussion}

The analysis of code descriptions together with the qualitative analysis of codes with low diversity and high exact match count suggests that parts of the ICD coding task can be performed in a simple and efficient manner, for example with rule-based systems. This opens up research potential for hybrid approaches. 

We have shown that match counts that take proximity into account are able to capture nuances of extracted evidence that would otherwise be missed with other metrics. 

The findings, that models are more likely to be wrong when fewer words are extracted and that duplicates are frequent, are counterintuitive (Occam's razor). As we would have expected more variance across models, further research is needed to investigate the precise role of robustness strategies on feature sparsity and associated model performance.

Considering the probabilities, there was no pattern that exact matches have higher output probabilities. A promising avenue may be to investigate which effect model calibration has on match types.

\subsection{Recommendations}

\paragraph{For practitioners}
Our findings demonstrate that the supervised approach has more exact matches, fewer empty, and fewer non-matches. 
If the implementation of a system requires high exact matches, our experiments show that current supervised methods are recommended. Investing in evidence annotation has a positive effect on addressing the requirement. For cases where exact matches have lower priority, the investment may not be necessary.
While the unsupervised training strategy with input gradient regularization proved similarly successful as the supervised approach, the annotated evidence was still used to compute a feature attribution threshold \cite{edin2024}. This is important when aligning human preferences about evidence with the evidence extracted from a model. 
To tackle duplicates and uppercasing, pre- and postprocessing steps can be helpful. Since explanations are on a token level, e.g., `pirin' as in `aspirin', they may be incomprehensible in isolation. For evidence presentation, it can be helpful to highlight entire words.  

\paragraph{For researchers}
Because of the common documents, it is not recommended to simply use Inpatient and Profee in a train-test scenario together because it can lead to data leakage. 

Furthermore, it is important to report and (if possible) evaluate false positives. 
Prior work used true positives and false negatives to evaluate explanations, that is, including codes not predicted by the model.  
False positives cannot be readily evaluated because of missing ground truth. 
Including false negatives is useful for benchmarking and finding suitable explanation methods. However, in a down-stream task, only predicted codes matter, whether correct or not.   
At a minimum, the number of false positives, which are not considered for the plausibility analysis, need to be reported. 
\section{Related Work}

\paragraph{ICD Coding}
The task of assigning medical codes given clinical notes is typically modeled as a multi-label classification problem. 
For automated ICD coding, two main approaches have emerged: (i) contextual approaches, often relying on CNNs or LSTMs, and (ii) pre-trained language models. 
The state-of-the-art for contextual approaches is CoRelation \cite{luo-etal-2024-corelation} which integrates a graph-based component. 
The predominant model architecture using pre-trained language models is PLM-ICD \cite{huang-etal-2022-plm} which shows similar performance to CoRelation. 
Regarding the evaluation procedure, \citet{edin2023critical} recently identified flaws in several works, e.g. inadequate train-test splits. 
The approaches are all comparable in performance, struggling with rare codes. 
Recent work has employed LLMs to perform ICD coding \cite{yang2023surpassinggpt4medicalcoding} but the performance is lagging behind the other approaches \cite{soroush2024}. 

\paragraph{Explainability and evidence extraction in ICD coding}
This research direction addresses the question of why a code is applicable to a clinical text. 
Here, the focus is on feature attribution methods which assign scores to the input features for each class. 
Several studies have utilized attention or explainability methods and applied them to medical coding problems \cite{mullenbach-etal-2018-explainable, ivankay-etal-2023-dare,hou2024, edin2024}. Until recently, there were few to no datasets with word-level evidence which led to evaluation concentrating on faithfulness \cite{wood-doughty-etal-2022-model} or unsustainable plausibility evaluation: experts rating informativeness \cite{mullenbach-etal-2018-explainable, kim-etal-2022-current}, 
approximating expert ratings \cite{wood-doughty-etal-2022-model} or anecdotal examples \cite{liu2022hierarchical}. 
With the MDACE dataset, plausibility can now be more readily evaluated using annotated evidence spans \cite{cheng-etal-2023-mdace} and this study focuses on plausibility evaluation. 
Perturbation-based methods, in particular LIME \cite{ribeiro2016} and KernelSHAP \cite{lundberg2017},  are not suitable for ICD coding due to high resource consumption. Instead, gradient-based methods are prevailing \cite{edin2024}.

\section{Conclusion}
In this work, we studied the task of explainable ICD coding based on medical records. We investigated the MDACE dataset which contains medical codes with corresponding evidence spans.  Our analysis revealed a certain overlap of evidence with code descriptions. 
We examined the performance of current approaches, comparing unsupervised and supervised methods, and found that overlap with the ground truth is relatively high. The supervised approach is better in extracting exact matches. 
Based on our findings, we proposed recommendations for extracting and evaluating evidence for ICD coding including the intended use of evidence and reporting of false positives. 
Going forward, there are several open research avenues in improving evaluation of semantic similarity, investigating the relation of explanation characteristics and performance, and a formalization of sufficient and complete evidence. 

\section*{Limitations}
One limitation is the amount of data available, which makes detailed error analyses challenging. Additionally, as pointed out in prior work, the data basis are discharge summaries, which may not fully represent real-world settings. 
We have decided on certain evaluation settings, such as a fixed context window where the value depends on the intended use of the explanations. We assume that explanations help guide attention to relevant information in the text for decision-making, but this may vary for other applications. Hence, our findings may not automatically generalize to other tasks or domains and should be specifically evaluated. 
Furthermore, we selected existing models based on explanation performance, but since explanation performance is inherently a secondary objective, in practice, models will likely be chosen based on classification performance.
While we provide considerations for evaluating semantically meaningful evidence, this remains a challenge. 
Our work inherits the limitations present in the MDACE and the underlying MIMIC data, such as annotation inconsistencies.

\section*{Ethical Considerations}
With the adoption of automated medical coding systems in applied settings there is a risk of overlooking errors which is why detailed analyses of model output is important. 
Medical coding systems offer to increase efficiency and reduce cost. Hereby it is not the goal to replace workers, but to assist coders in a collaborative human-AI effort to solve the task of ICD coding. Explanations should support decision making. However, explanations can lead to undesired effects, such as over-interpretation, over-trust or algorithm aversion. It is important to focus on evaluation of explanations to ensure that they are comprehensible and relevant before blindly integrating them in an application.

\section*{Acknowledgments}
The work of DA and ES was done within the SmartHospital.NRW project, funded by the Ministry of Economic Affairs, Industry, Climate Action and Energy of the State of North Rhine-Westphalia, Germany.
We thank the reviewers for their valuable feedback.

\bibliography{custom}

\appendix

\section{Details on Problem Setting}
Medical documents have multiple associated ICD codes and code prediction is therefore treated as a multi-label classification task. 
Given a set of documents \( X \), the objective is to learn a function \( f \) that maps each document \( x_i \in X \) to a probability vector \( \hat{y} \in [0, 1]^N \), where \( N \) is the set of codes. 
Given a clinical text with a token sequence, the classification objective is to determine the output probabilities of the set of codes. 
The secondary objective is evidence extraction, here, approached with feature attribution methods. An attribution function assigns a score to each token that reflects the token’s influence on the model’s prediction. 

\section{Data Details}
The MDACE data is built using a subset of MIMIC-III data \cite{johnson2016mimic}. The data includes hospital records of patients that holds information on demographic data, diagnoses and procedures. Each hospital admission has several associated documents which were all coded using ICD systems. 
The ICD codes are especially useful for billing purposes. 
While the MIMIC-III dataset contains labels on a document-level, MDACE additionally provides evidence for each of these labeled codes on a textual level. 
The data is provided in a json format. 
Each note contains an ID, document category, and a list of annotations. An annotation contains the ICD code, the code description, and, most importantly, begin and end position which refer to character position in the note text string.
There can be several annotations in a text with the same ICD code, as in the case of Profee annotations. 

\section{Implementation Details}
\label{sec:appendix}

\subsection{Seed selection}
\begin{figure}
    \centering
    \begin{subfigure}{0.48\textwidth}
    \hspace*{-7pt}\input{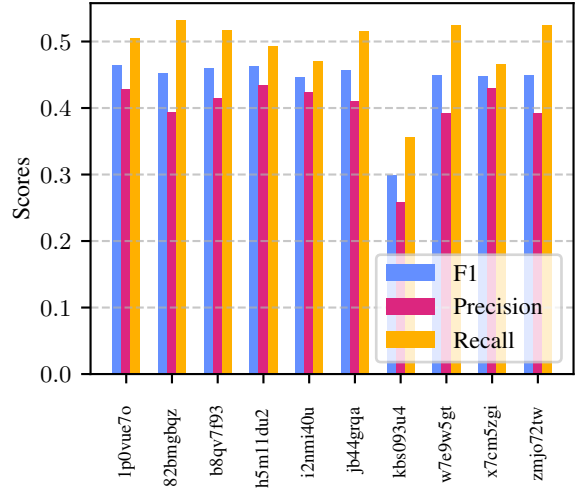}
    \caption{IGR}
    \end{subfigure}
    \begin{subfigure}{0.48\textwidth}
    \hspace*{-7pt}\input{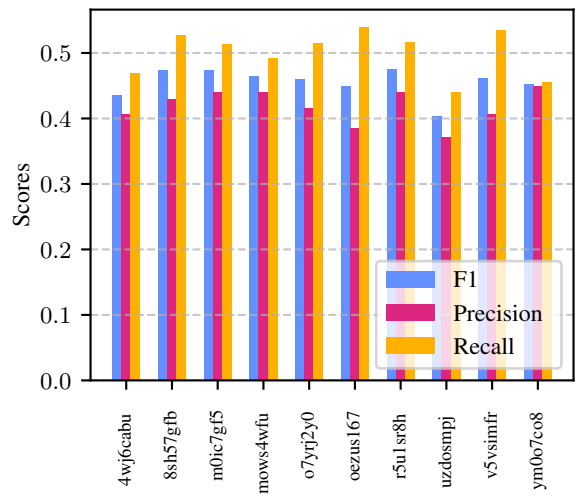}
    \caption{Supervised}
    \end{subfigure}
    \caption{F1 scores, precision and recall in regards to explanations for all seeds of supervised and IGR \cite{edin2024}. Top: IGR, bottom: supervised}
    \label{fig:all_seeds}
\end{figure}

Several research questions focused on two approaches, supervised training and input gradient regularization (IGR) for which one seed was selected respectively. 
The performance of evidence extraction for all seeds of these approaches is depicted in \autoref{fig:all_seeds}.

\subsection{Overlap between evidence and code descriptions}

The lemmatization, stop word and punctuation removal is performed with spaCy \cite{Honnibal_spaCy_Industrial-strength_Natural_2020}.
Numbers are not removed. 
With the cleaning process `Shigellosis due to Shigella flexneri' is reduced to `Shigellosis Shigella flexneri', while for `Amebiasis, unspecified' only the comma is removed.
The stopword removal also includes words such as `without' and `except' where the removal may be undesired. 

\end{document}